\documentclass{article}

\usepackage{arxiv}

\usepackage[utf8]{inputenc} 
\usepackage[T1]{fontenc}    
\usepackage{hyperref}       
\usepackage{url}            
\usepackage{booktabs}       
\usepackage{amsfonts}       
\usepackage{nicefrac}       
\usepackage{microtype}      
\usepackage{lipsum}		
\usepackage{graphicx}
\usepackage{natbib}
\usepackage{doi}
\usepackage{makecell}
\usepackage{pifont}
\usepackage{amsmath}
\usepackage{multirow}

\title{CompleteDT: Point Cloud Completion with Dense Augment Inference Transformers}

\author{
	Jun Li, Shangwei Guo, Shaokun Han*\\
	Beijing Key Lab for Precision Optoelectronic Measurement Instrument and Technology, Beijing Institute of Technology\\
	School of Optics and Photonics, Beijing Institute of Technology\\
	Beijing, 100081, China\\
	\texttt{skhan@bit.edu.cn}(* indicates the corresponding author)
}

\renewcommand{\shorttitle}{\textit{arXiv} Template}

\hypersetup{
pdftitle={A template for the arxiv style},
pdfsubject={q-bio.NC, q-bio.QM},
pdfauthor={David S.~Hippocampus, Elias D.~Striatum},
pdfkeywords={First keyword, Second keyword, More},
}

\begin{document}
\maketitle

\begin{abstract}
Point cloud completion task aims to predict the missing part of incomplete point clouds and generate complete point clouds with details. In this paper, we propose a novel point cloud completion network, namely CompleteDT. Specifically, features are learned from point clouds with different resolutions, which is sampled from the incomplete input, and are converted to a series of \textit{spots} based on the geometrical structure. Then, the Dense Relation Augment Module (DRA) based on the transformer is proposed to learn features within \textit{spots} and consider the correlation among these \textit{spots}. The DRA consists of Point Local-Attention Module (PLA) and Point Dense Multi-Scale Attention Module (PDMA), where the PLA captures the local information within the local \textit{spots} by adaptively measuring weights of neighbors and the PDMA exploits the global relationship between these \textit{spots} in a multi-scale densely connected manner. Lastly, the complete shape is predicted from \textit{spots} by the Multi-resolution Point Fusion Module (MPF), which gradually generates complete point clouds from \textit{spots}, and updates \textit{spots} based on these generated point clouds. Experimental results show that, because the DRA based on the transformer can learn the expressive features from the incomplete input and the MPF can fully explore these feature to predict the complete input, our method largely outperforms the state-of-the-art methods.
\end{abstract}

\section{Introduction}\label{sec:Introduction}

Point cloud data is one of the most popular 3D shape representations to express the attributes of real-world objects. 3D sensors such as LiDAR and Kinect are widespread used in point cloud acquisition process. However, the main limitations of these sensors are their resolutions and viewing angles, so geometric information may be not accurately acquired or even lost, which leads to incomplete and sparse point clouds\citep{huang2020pf}. Therefore, of particular concern is the 3D shape completion by learning the potential features of incomplete point clouds and then generating complete point clouds. The task of point cloud completion contributes to downstream 3D computer vision (CV) tasks like classification\citep{qi2017pointnet,qi2017pointnet++}, segmentation\citep{lei2020seggcn,xu2020weakly} and target detection\citep{ali2018yolo3d,chen2017multi}. 

In recent years, attention has been focused on point cloud completion tasks, which predict the missing shapes and maintain fine-grained complete shapes. The main concept is to learn the geometric properties and apparent structures to complete the point cloud. Some works\citep{dai2017shape, yang20173d} propose to perform the 3D convolution on voxels by voxelizing the point cloud. However, with the increase of the voxel resolution, these methods suffer from a heavy computational cost. Compared to the above methods voxelizing the point cloud, an increasing number of methods\citep{mandikal2019dense,yuan2018pcn,achlioptas2018learning,groueix2018papier,sarmad2019rl} based on PointNet\citep{qi2017pointnet} and PointNet++\citep{qi2017pointnet++} directly process 3D coordinates to complete the point cloud, which greatly save the computational cost. Point Cloud Completion (PCN)\citep{yuan2018pcn}, as the pioneer work based on PointNet\citep{qi2017pointnet}, adopts an encoder-decoder framework to decode a complete shape from global features extracted by the encoder. However, PCN\citep{yuan2018pcn} has failed to benefit from local features, and can’t generate fine geometric details. Other methods\citep{wang2020cascaded,wen2020point,wen2021pmp} explore the usefulness of the local information to improve the performance of point cloud completion. Local features can express the structure information by relating points and their neighborhood points. For example, SA-Net\citep{wen2020point} exploits the local details by a skip-attention mechanism, which can selectively emphasize geometric information from the local regions for the generation of complete point clouds. 

The current methods\citep{wang2020cascaded,pan2020ecg} imitate the visual 2D Convolutional Neural Network (2D-CNN) to build a hierarchical network, where each layer extracts local features from neighbors.  As the network deepens and the receptive field increases, it aggregate local features to global features. Because of the imitation of 2D-CNN, these methods inherit drawback of 2D-CNN, that is, the lack of ability to extract global features. The global features can express the information of the entire point cloud, and can provide important semantic contexts. Thus, it is important to use global features in the procession of point cloud completion. However, the transformer\citep{vaswani2017attention} aggregates information from the entire point cloud to output features, so it has a global receptive field to extract the global features of the point cloud. To learn local detail features while exploring global features, hence improving the performance of point cloud completion, we propose a network based on the transformer\citep{vaswani2017attention}, named CompleteDT.  Our model is structured in four components: \textbf{1) ResMLP Module:} ResMLP Module extracts features by performing linear transformation at point level. To fully extract features from the point cloud to complete the incomplete input, multi-resolution feature maps are obtained from these features with Iterative Farthest Point Sampling (IFPS). \textbf{2) Self-Guided Relation Module (SGR):} The local information of point clouds plays an important role for point cloud completion. The SGR is designed to aggregate local information from multi-resolution feature maps and to sequentially process these feature maps from low to high resolution to generate \textit{spots} with multiple resolutions, which contain local information and semantic contexts. \textbf{3) Dense Relation Augment Module (DRA):} To further extract the local information and explore global relationships between \textit{spots}, we design the DRA, which is based on the transformer\citep{vaswani2017attention}. This module is composed of two key components, i.e., Point Local Attention Module (PLA) and Point Dense Multi-Scale Attention Module (PDMA). PLA adaptively learns local features only from neighbors around \textit{spots}. PDMA can densely extract the global correlation among all \textit{spots}. It divides the multi-head attention into different groups. Each group receives the set of the outputs of all previous groups and aggregates them to a representation with an embedding dimension. This design makes the structure of PDMA quite different from that of the transformer\citep{vaswani2017attention}. DRA adopts PLA and PDMA in parallel to extract the global information and retain local features. \textbf{4) Multi-resolution Point Fusion Module (MPF):} We devise a multi-resolution generation module for completing point clouds in a coarse-to-fine manner. MPF first generates \textit{global spots}, which represent the information about the complete shape, and then uses \textit{spots} at different semantic levels to progressively generate complete shapes with different resolutions and update \textit{global spots}. Finally, the updated \textit{global spots} can be used to recover the final complete point cloud.

Our main contributions can be summarized as follows:
\begin{itemize}
	\item We propose a novel module DRA consisting PLA and PDMA, which extracts adequate features containing local information and global semantics. PLA enhances to inference the local information of \textit{spots}. PDMA, where embedding dimensions of kernels of multi-head attention are different, extracts the global correlation among all \textit{spots}. Thus, DRA provides detailed local information and global correlations for the Multi-resolution Point Feature Fusion to generate a complete point cloud.
	
	\item We propose the MPF to generate a high-quality complete point cloud in the way that the spatial position of points control distribution of features. Specifically, it first generates a complete point cloud on an unorganized sets of 3D feature maps, and then updates the features based on the spatial structure of this point cloud, thereby forcing the geometric mapping of the point cloud onto the feature space. 
	
	\item We propose a novel network CompleteDT for point cloud completion task. CompleteDT can learn local features within each spot and explore the global correlation among \textit{spots}. Compared with previous methods of point cloud completion, CompleteDT achieves the state-of-art performance of point cloud completion.
\end{itemize}

\begin{figure*}[t]
	\centering
	\includegraphics[width = \linewidth]{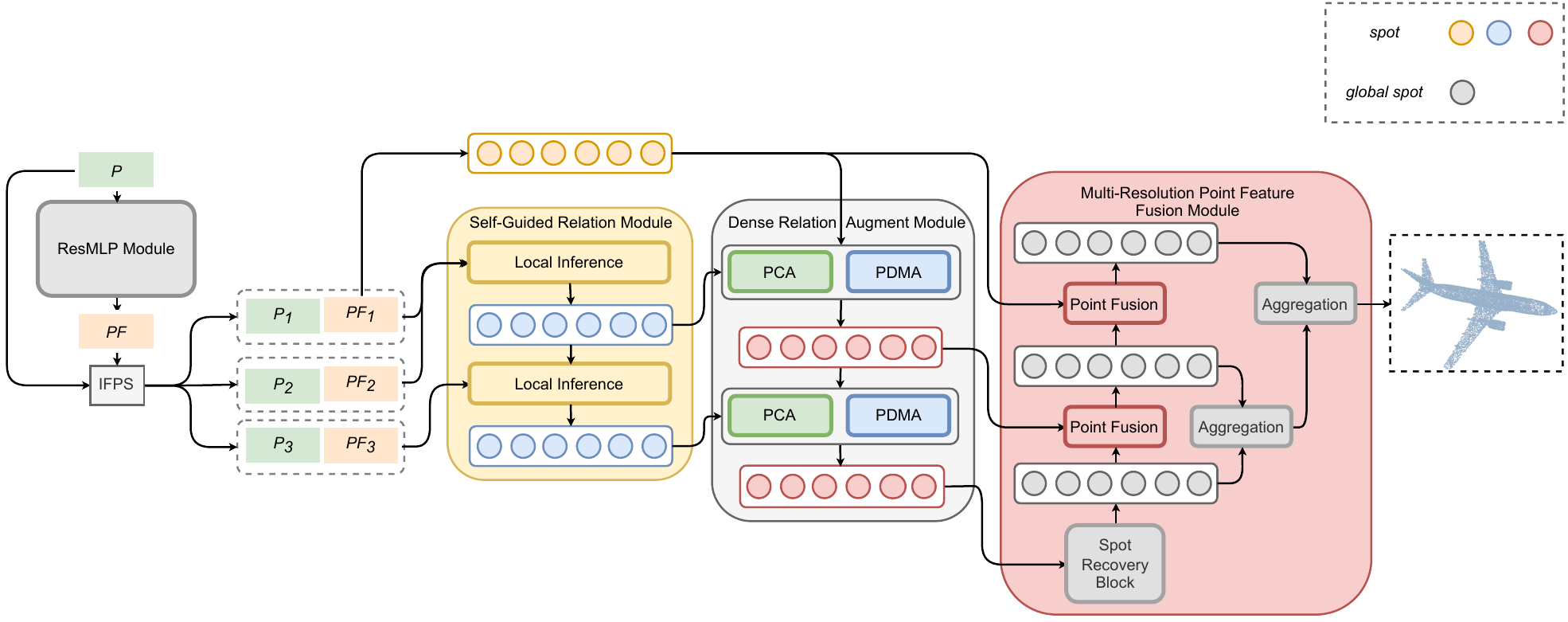}
	\caption{The overall architecture of CompleteDT, which consists of four modules: ResMLP Module, Self-Guided Relation Module, Dense Relation Augment Module and Multi-resolution Point Fusion Module. We first use ResMLP Module based on MLP to extract features. Then multi-resolution point clouds and their features are obtained. After obtaining \textit{spots} (blue circle) with Self-Guided Relation Module, we use a transformer-based module to enhance the inference of \textit{spots} (red circle). Finally, we obtain \textit{global spots} (gray circle) and gradually update them to generate the complete point cloud.}\label{fig:OVERALL}
\end{figure*}

\section{Related Work}\label{sec:Related Work}
\subsection{3D Shape Completion}\label{sec:Related Work/3D Shape Completion}
\textbf{Based on voxel data} Due to the unordered property of point clouds, some methods\citep{dai2017shape, yang20173d,wang2019forknet,han2017high,nguyen2016field,stutz2018learning,varley2017shape} voxelize point clouds into rigid grids and then perform 3D convolution on them. For example, the 3D-Encoder-Predictor Network (3D-EPN)\citep{dai2017shape} generates a low-resolution complete output, and then refines it from a shape database. Although the voxel-based method can complete the shape, its details are lost for the introduced quantization error, and its computational cost increases significantly with increasing resolution.

\textbf{Based on point cloud data} To solve the unordered structure of point cloud, PointNet\citep{qi2017pointnet} is proposed to analyze point clouds on point level and has benefited to many downstream tasks. For point cloud completion, PCN\citep{yuan2018pcn}, as the pioneer work based on PointNet\citep{qi2017pointnet} for completing the point cloud, extracts the global representation from the incomplete point cloud and then decodes this representation to generate the point cloud from coarse to detail. However, the point cloud generated by PCN\citep{yuan2018pcn} lacks details due to ignoring local information. After PCN\citep{yuan2018pcn}, other methods emerge and recover the point cloud with the higher resolution. For example, PF-Net\citep{huang2020pf} is an encoder-decoder network based on the Adversarial Generative Networks (GAN). PF-Net\citep{huang2020pf} suggests to predict the missing shape of the input instead of generating the overall shape, and then forms the complete shape with the missing shape and the input. PF-Net\citep{huang2020pf} can well retain the spatial arrangements of the input and figure out the detailed geometrical structure in missing. VRCNet\citep{pan2021variational} is the first network for point cloud completion based on the probabilistic model. VRCNet\citep{pan2021variational} first predicts the coarse shape with “probabilistic modeling” (PMNet) and then recovers shape details with “relational enhancement” (RENet). GRNet\citep{xie2020grnet} refocus on the voxel data for its rigid structure. GRNet\citep{xie2020grnet} introduces 3D grids as an intermediate representation to generate the complete point cloud, thereby recovering the structural and context information of the point cloud. SoftPoolNet\citep{wang2020softpoolnet} generates high-resolution complete point cloud by sorting features into a feature map to obtain local groups and then applying the 2D convolution. Based on the SoftPoolNet\citep{wang2020softpoolnet}, SoftPool++\citep{wang2022softpool++}, as a feature extraction module, can be applied multiple times in an encoder-decoder architecture by the use of point-wise skip connections. Based on the above design, model based on SoftPool++\citep{wang2022softpool++} can generate complete point clouds that preserve geometric details.

\subsection{Transformer on 3D Point Cloud}\label{sec:Related Work/Transformer on 3D Point Cloud}
Transformers\citep{vaswani2017attention} are first applied in the field of Natural Language Processing (NLP), and process data in parallel and order-independent. Transformers\citep{vaswani2017attention} can extract long-range features due to its capability of aggregating global features. The subsequent success\citep{devlin2018bert,liu2021swin,rao2021dynamicvit} of transformers\citep{vaswani2017attention} in computer vision (CV) greatly aroused transformer's interest in the field of 3D point clouds. Based on the transformer\citep{vaswani2017attention}, a new learning framework PCT\citep{guo2021pct} is proposed. PCT\citep{guo2021pct} first extracts local features from grouped points and then inputs these local features to the module based on the transformer\citep{vaswani2017attention} for learning the global representation. In the realm of point cloud completion, PMP-Net++\citep{wen2022pmp++}, introduces a self-attention transformer\citep{vaswani2017attention} module between each set abstraction (SA) layer of PointNet++\citep{qi2017pointnet++} based encoder, strictly learns the correspondence at the point level and moves each point of the incomplete input to improve the quality of the complete point cloud. CompleteDT benefits from transformer\citep{vaswani2017attention} to augmenting the inference of features. One striking characteristic of our approach is that we use a transformer based on a self-attention mechanism with kernels embedding different dimensions, which can ensure the effective fusion of features at semantic information. PoinTr\citep{yu2021pointr} converts the point cloud to a sequence of point proxies and devises a transformer-based block that completes the shape. PoinTr\citep{yu2021pointr} first exploits the self-attention mechanism to explore long-range semantic correlations, then uses a linear layer to learn local geometric information on local proxies captured by K-nearest neighbor (KNN). Finally, these two kinds of results are fused to ensure global information and local information, and recover detailed information for point cloud completion. The difference exists between PoinTr\citep{yu2021pointr} and CompleteDT is that CompleteDT is concerned with the performance of transformers\citep{vaswani2017attention} on local information and designs the PLA (Section~\ref{sec:Methods/DRA}) to learn the local information, while PoinTr\citep{yu2021pointr} focuses on the description of local information with KNN. CompleteDT adaptively measures neighborhood weights to better infer local information. The ablation study (Section~\ref{sec:Ablation Study/DRA}) proves the efficiency of the PLA. 

\section{Methods}\label{sec:Methods}
In this section, we first introduce the proposed model CompleteDT, which can learn useful features from incomplete point clouds and then infer complete shapes. Next, we show the mentioned modules separately. Finally, we present loss functions.

\subsection{Model}\label{sec:Methods/Model}
Our goal is to complete the point cloud with more details by extracting the features within neighborhoods of the point cloud and exploring the interaction among these neighborhoods. It may be insufficient to only integrate local features while ignoring the global relationship among neighborhoods\citep{cui2021geometric}. However, the difficulty in the extraction of global relationship exists because local features within receptive fields is extracted layer by layer and then global information is aggregated from these local features as the network deepens and receptive fields increases. The recent success in the transformer\citep{vaswani2017attention} has clearly demonstrated the effectiveness of resolving the disorder of point clouds and extracting long-range features. Thus, we propose a network based on the transformer\citep{vaswani2017attention} for point cloud completion.

As shown in Figure~\ref{fig:OVERALL}, we first use the ResMLP Module as preprocess (Section~\ref{sec:Methods/ResMLP}) to extract features from the input and then obtain multi-resolution point clouds and multi-resolution feature maps with IFPS. Then, we design the Self-Guided Relation Module (SGR) (Section~\ref{sec:Methods/SGR}) to obtain the \textit{spots} by aggregating local information from multi-resolution feature maps and then sequentially processing these feature maps from low to high resolution. To be further infer the \textit{spots} to improve the performance of point cloud completion, we present the Dense Relation Augment Module (DRA) (Section~\ref{sec:Methods/DRA}) consisting of the Point Local Attention Module (PLA) and the Point Dense Multi-Scale Attention Module (PDMA). The PLA can enhance the inference of features within the \textit{spots}. The PDMA considers information of \textit{spots} from different resolution point clouds comprehensively. Concretely, the \textit{spots} from high-resolution point cloud contain more detailed information, while \textit{spots} from low-resolution point cloud tends to retain the knowledge of overall geometry. Therefore, outputs of two kinds of modules are fused to obtain the representation with both long-range and short-range information. Based on expressive features learned from DRA, we design the Multi-resolution Point Fusion Module (MPF) (Section~\ref{sec:Methods/MPF}) to generate a high-resolution complete shape. Specifically, it first generates \textit{global spots} containing global information of a complete shape, then obtains multi-resolution point clouds and updates \textit{global spots} based on the spatial structure of these point clouds. Spatial positions of point clouds is forced to map onto the distribution of \textit{global spots}. The updated \textit{global spots} can be used to generate the final complete shape with the high resolution.

\subsection{ResMLP Module}\label{sec:Methods/ResMLP}
\begin{figure}[h]
	\centering
	\includegraphics[width = 0.8\linewidth]{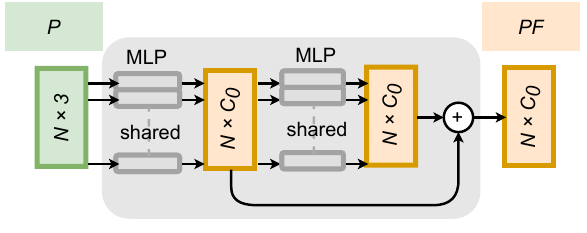}
	\caption{Illustration of the ResMLP Module, which mainly consists of two MLP layers.}\label{fig:ResMLP}
\end{figure}
As one of the most well-known methods for point cloud analysis, PointNet\citep{qi2017pointnet} proposes to directly process 3D coordinates of points with multi-layer perceptron layers (MLP). MLP facilitates the transformation of features for its simple nature and the powerful representation capability\citep{xie2020grnet}. Inspired by this idea, we present a module named ResMLP Module based on the residual mechanism. ResMLP is a pre-processor to extract features on each point, and the architecture of it is shown in Figure~\ref{fig:ResMLP}. Specifically, given a point cloud $P\in \mathbb{R}^{N\times3}$, we perform two MLP layers sequentially on this point cloud $P$ and then sum these results in pairwise with the skip connection as the final output $PF\in \mathbb{R}^{N\times C_0}\ (C_0=64)$. 

Due to the shape of $P$ is incomplete, features can only be inferred from the visible shape of $P$. To extract as much geometry knowledge as possible from $P$, we propose to sample point clouds\citep{qi2017pointnet++} to get multi-resolution feature maps. Features of a high-resolution point cloud can provide detail information. Although features of a low-resolution point cloud may ignore the details, they are closer to the overall skeleton features. As shown in Figure~\ref{fig:OVERALL}, we obtain multi-resolution point clouds $P_m\in \mathbb{R}^{N_m\times3}\ (m=1,2,3,N_1>N_2>N_3)$ and feature maps $PF_m\in \mathbb{R}^{N_m\times C_0}$ by sampling $P$ and $PF$ with IFPS. 

\subsection{Self-Guided Relation Module}\label{sec:Methods/SGR}

\begin{figure}[h]%
	\centering
	\includegraphics[width = \linewidth]{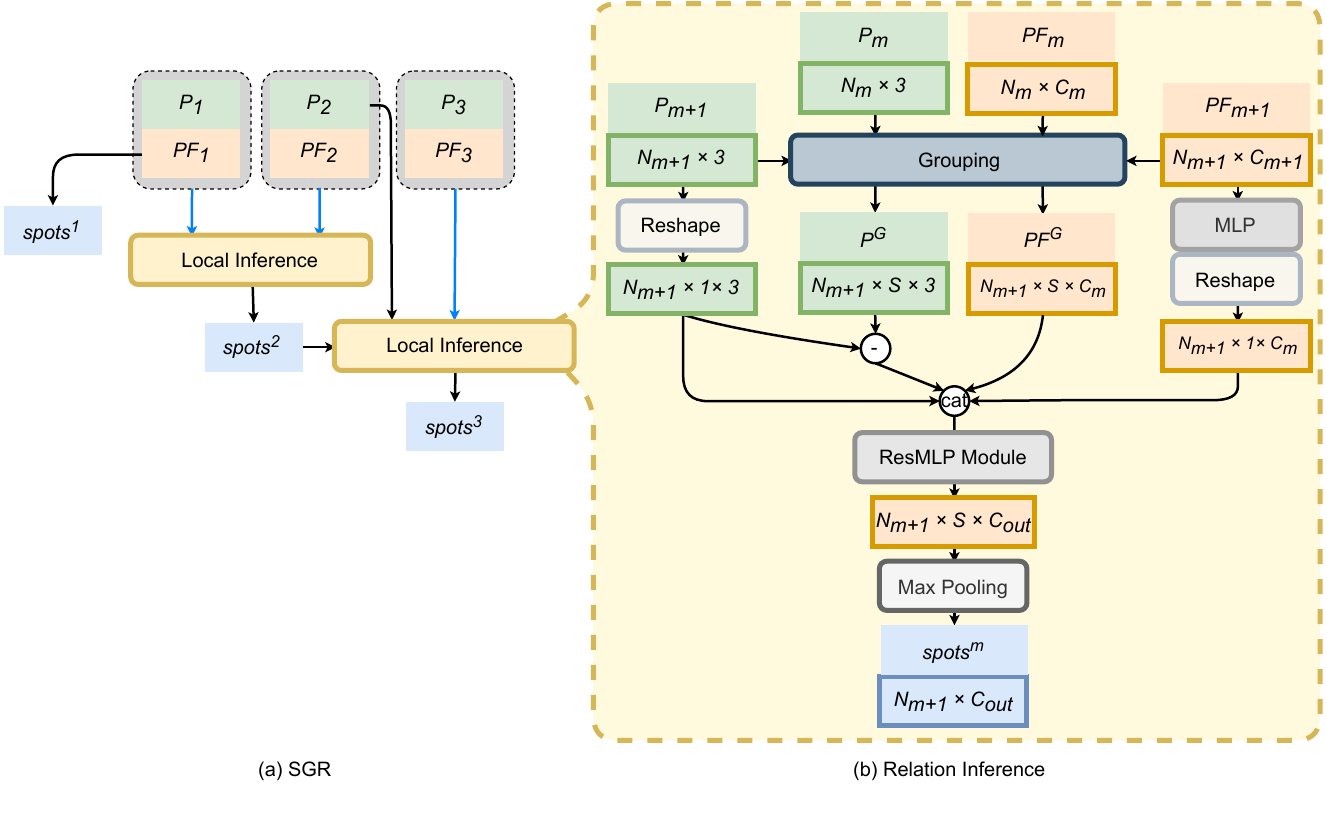}
	\caption{Illustration of the Self-Guided Relation Module, which is responsible for generating $spots^1$,  $spots^2$ and  $spots^3$. $spots^1$ indicates  features $PF_1$, $spots^2$ and $spots^3$ mean features obtained by Local Inference. The blue arrow indicates that the set of $\{P_m,PF_m\}$ inputs to Local Inference. The superscript ``G" means ``Group"}\label{fig:SGR}
\end{figure}

To take full advantage of the multi-resolution point clouds $P_m$ and their feature maps $PF_m$, we design the Self-Guided Relation Module (SGR) for gradually fusing local features in a self-guided way. SGR uses the hierarchical relationship of point clouds with different resolutions to sequentially aggregates features. Considering the hierarchical relationship ensures that the high-resolution information provides local details for the low-resolution information, while the features of the low resolution guide the integration of high-resolution features. As a result, semantic information is enhanced due to features being aggregated in a hierarchical manner. Therefore, it is effective for fusing features to consider hierarchical relationships. 

For two point clouds with different resolution, the local information of a point in the lower one can be provided by the higher one. This point in the low-resolution point cloud and its local features in the high-resolution point cloud can be defined as a \textit{spot}. As shown in Figure~\ref{fig:SGR} (a), we consider the output of $m_{th}$ stage of SGR as $spots^m$.  $PF_{2}$ guides the fusion of $PF_1$ to obtain $spots^2$ with Local Inference, which is an efficient block of SGR. Then the $spots^2$ propagates its information to $PF_{3}$ to generate $spots^3$. Especially, $spots^1$ indicates features $PF_1$. 

Figure~\ref{fig:SGR} (b) shows the details of Local Inference. Taking any two sets of $\{P_m,PF_m\}$ and $\{P_{m+1},PF_{m+1}\}$ as an example, we implement grouping operation proposed by PointNet++\citep{qi2017pointnet++} to capture local features. Simply, for $p_i\in P_{m+1}$, we first gather its neighborhoods $N_K$ (Eq.~\ref{eq:K}), and then obtain its neighbor points $P^G\in P_m$ and their features $PF^G\in PF_m$ (Eq.~\ref{eq:PPF}).  

\begin{figure*}[h]%
	\centering
	\includegraphics[width = 0.9 \linewidth]{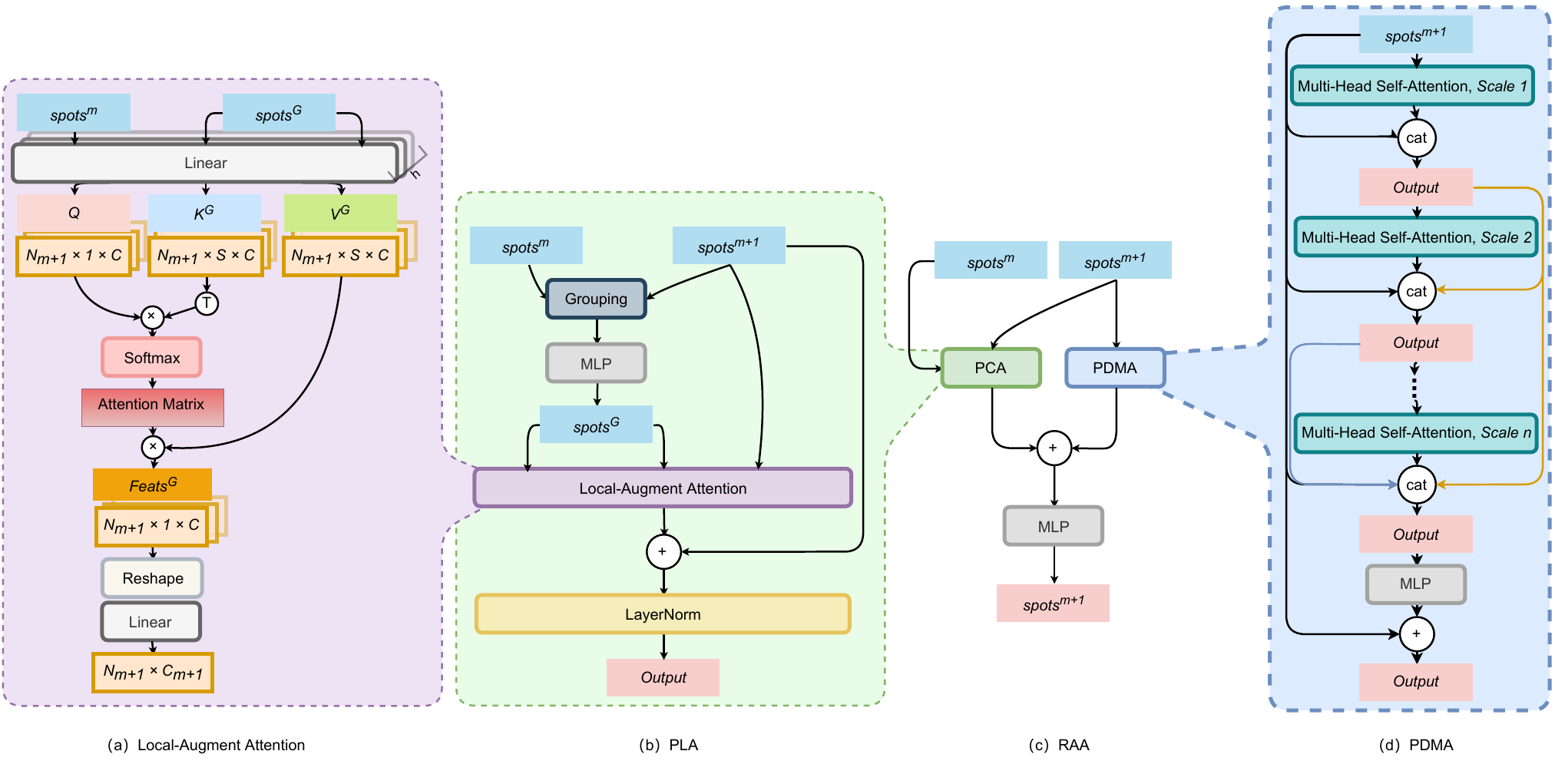}
	\caption{Illustration of the $m_{th}$ stage of Dense Relation Augment Module. This module consists of PLA (b) and PDMA (d). PLA is composed of the Local-Augment Attention (a).}\label{fig:DRA}
\end{figure*}

\begin{equation}
	\small
	\label{eq:K}
	N_K\left(p_i,P_m\right)=\{k\mid d\left(p_i,p_k\right)<r, p_k\in P_m\}
\end{equation}
where, $d\left(\cdot\right)$ is a distance function and $r$ is the threshold.
\begin{equation}
	\begin{aligned}
		\label{eq:PPF}
		P^{G}&=\bigcup_{k\in N_K(p_i,P_m)}{p_k\in P_m}\\
		PF^{G}&=\bigcup_{k\in N_K(p_i,P_m)}{pf_k\in PF_m}
	\end{aligned}
\end{equation}

In addition to being the input of grouping operation, $PF_{m+1}$ also needs to map its dimensions to $PF_m$. Then, we obtain the relative relationship by subtracting between $P_m$ and $P^G$ in the channel dimension. After concatenating these representations on the channel dimension, we obtain $spots^{m+1}$ from two sets of $\{P_m,PF_m\}$ and $\{P_{m+1},PF_{m+1}\}$ by using the ResMLP Module (Section~\ref{sec:Methods/ResMLP}) and the Max-pool layer. Therefore, $spots^{m+1}$ contain the details of $PF_{m}$ and the semantic features of $PF_{m+1}$. Then $spots^{m+1}$ are used to aggregate its own information into $PF_{m+2}$ to generate $spots^{m+2}$. In this way, SGR aggregates features sequentially based on hierarchical relationships and generates \textit{spots} containing local information.
\subsection{Dense Relation Augment Module}\label{sec:Methods/DRA}

The $spots^m$ generated from SGR(Section~\ref{sec:Methods/SGR}) contain local information. It may be insufficient to consider local information. The global correlation between any two $spot_i, spot_j\in spots^m\ (i\neq j)$ should be taken into account. However, the common operation like convolution is restrictive in acquiring global correlation. The transformer\citep{vaswani2017attention} contributes to solve this problem for its ability of obtaining long-range information for the processing of any feature. Based on the transformer, we design a valid module, named Dense Relation Augment Module (DRA). In this section, we first review the transformer\citep{vaswani2017attention}, then introduce DRA and its important components.

\subsubsection{Review of the transformer}
One interpretation of transformers applied to point clouds is to regard the point cloud as a sentence and its point as words\citep{guo2021pct}. The common form of transformer is based on self-attention mechanism. Given the input $F\in \mathbb{R}^{N\times C_F}$, it is performed by a the linear transformations with kernel weights $W_q,W_k,W_v$ to finish the self-attention mechanism. We use the symbol $\mathcal{H}$ to represent the operation of self-attention mechanism:
\begin{equation}
	\small
	\label{eq:vanillaSA}
	f=\mathcal{H}(F,W_q,W_k,W_v)
\end{equation}
where $W_q\in \mathbb{R}^{C_F\times C}$, $W_k\in \mathbb{R}^{C_F\times C}$ and $W_v\in\mathbb{R}^{C_F\times C}$ are shared learnable parameters

$F$ is projected into the query $Q\in \mathbb{R}^{N\times C}$, key $K\in \mathbb{R}^{N\times C}$ and value $V\in \mathbb{R}^{N\times C}$with the linear transformations\citep{vaswani2017attention}: 
\begin{equation}
	\label{eq:vanillaQKV}
	Q,K,V=F\otimes W_q,F\otimes W_k,F\otimes W_v
\end{equation}
where ``$\otimes$" represents the matrix multiplication.

The specific operation process is shown in Figure~\ref{fig:attention} (a). The attention matrix $A\in \mathbb{R}^{N\times N}$ is generated by performing the matrix multiplication on $Q$ and $K$, normalizing and applying softmax function sequentially. The $a_i^j\in A$ can be expressed as Eq.~\ref{eq:vanillaA}.

\begin{equation}
	\small
	\label{eq:vanillaA}
	a_i^j=softmax(\frac{\sum\limits_{c=1}^C q_i^ck_j^c}{\sqrt{d_{s}}})
\end{equation}
where $q_i\in Q$, $k_j\in K$, and $d_s$ is the value of normalization.

The attention matrix $A$ is multiplied by $V$ to obtain $Feats$. The feature $f_i^c\in Feats$ can be expressed as follows:
\begin{equation}
	\label{eq:vanillaF}
	f_i^c=\sum_{n=1}^{N}{\alpha_i^nv_n^{c}}
\end{equation}
where $v_n^{c}\in V$. The letter ``$C$" means the number of the channel dimension. The letter ``$N$" indicates the number of point cloud.

\subsubsection{The architecture of DRA}
Based on the self-attention mechanism, we design a valid module, named Dense Relation Augment Module (DRA). Our DRA consists of Point Local-Attention Module (PLA) and Point Dense Multi-Scale Attention Module (PDMA). DRA adopts the same feature aggregation method as SGR (Section~\ref{sec:Methods/SGR}), which tries to further infer local information of $spots$ based on hierarchical dependencies. For the convenience of explanation, we take $m_{th}$ stage of DRA as an example. For $spots^{m}$ and $spots^{m+1}$, DRA utilizes PLA and PDMA in parallel to boost the feature inference procession. As shown in Figure~\ref{fig:DRA} (c), PLA facilitates to extract the local information by fusing features of $spots^{m}$ into that of $spots^{m+1}$, and PDMA favors to obtain the global correlation among all elements of $spots^{m+1}$. Obtaining the output of DRA first sums the results of PLA and PDMA in pairwise and then applys the MLP layer. In this way, $spots^{m+1}$ are updated by DRA not only contains local information, but also supplements global dependencies. 

\subsubsection{The architecture of PLA}
PLA migrates the transformer’s property of learning long-range features to local features and measures weights of local information adaptively. As shown in Figure~\ref{fig:DRA} (b), PLA, including the Local-Augment Attention (Figure~\ref{fig:DRA} (a)), is an attention module based on residual block for strengthening the inference of local features. For the $spots^{m+1}\in R^{N_{m+1}\times C_{m+1}}$, we obtain $spots^G\in R^{N_{m+1}\times S\times C_m}$ by gathering its neighborhood $spots$ from $spots^m\in R^{N_m\times C_m}\left(N_m>N_{m+1}\right)$ based on neighborhoods $N_K$ (Eq.~\ref{eq:K}) from SGR and applying the MLP. After obtaining $spots^G$, it and $spots^{m+1}$ input to the Local-Augment Attention.
\begin{figure}[h]%
	\centering
	\includegraphics[width = 0.4\linewidth]{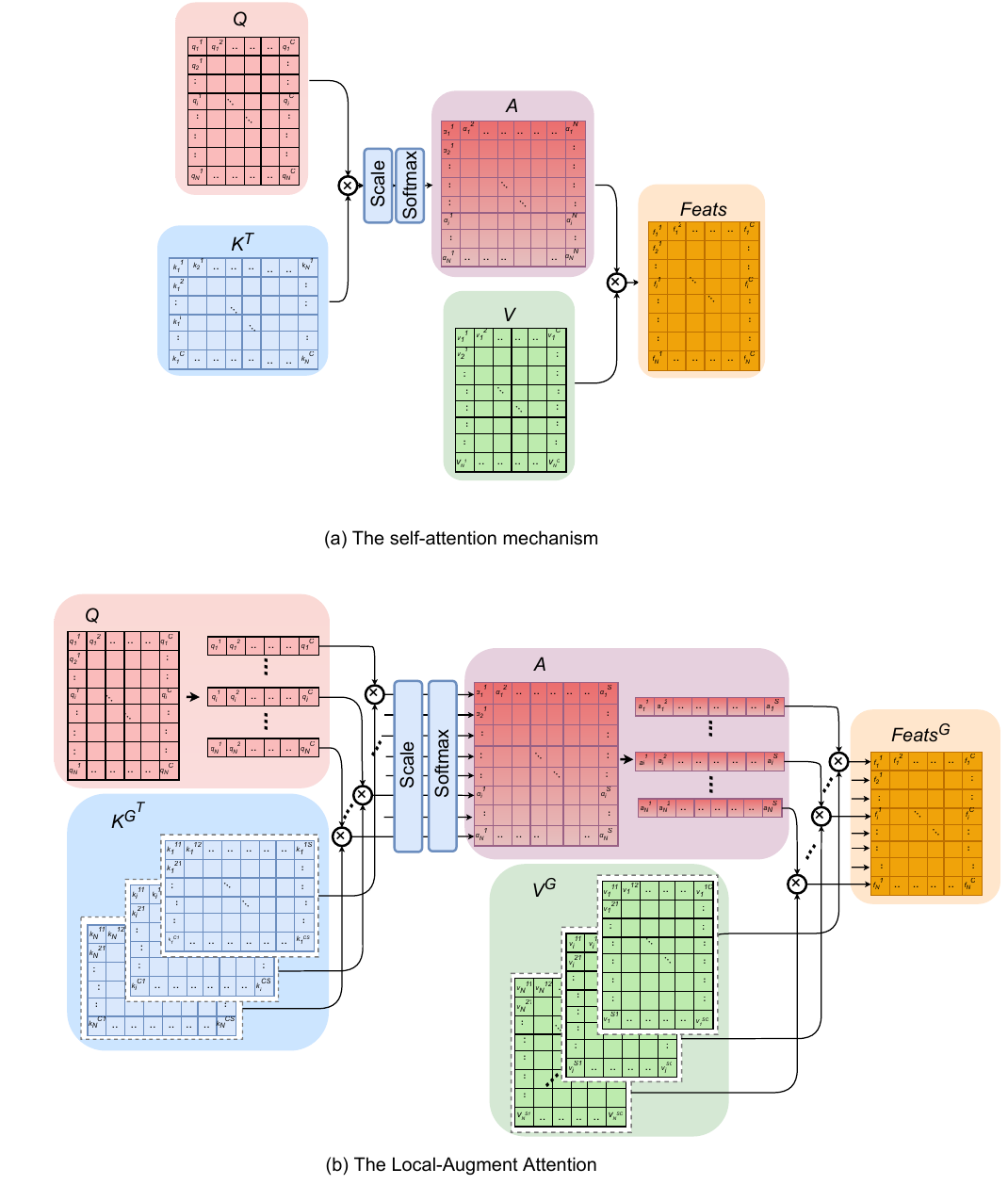}
	\caption{The specific implementation of the vanilla transformer and Local-Augment Attention. Scale means the normalized parameter.}\label{fig:attention}
\end{figure}

Different from the self-attention that obtains semantic relation among all features, Local Augment Attention uses the feature similarity to capture the geometric correlation between a $spot$ and its certain local \textit{spots}. The specific implementation of our Local Augment Attention is shown in the Figure~\ref{fig:attention} (b). $spots^{m+1}$ is used to obtain queries $Q\in \mathbb{R}^{N\times1\times C}$, and $spots^G\in spots^m$ is used to get keys $K^G\in \mathbb{R}^{N\times S\times C}$ and values $V^G\in \mathbb{R}^{N\times S\times C}$. Eq.~\ref{eq:vanillaQKV} becomes:
\begin{equation}
	\label{eq:PLAQKV}
	\begin{aligned}
		Q &= spots^{m+1}\otimes W_q\\
		K^G,V^G&=spots^{G}\otimes W_k,spots^{G}\otimes W_v
	\end{aligned}
\end{equation}
where $W_q\in \mathbb{R}^{C_{m+1}\times C}$, $W_k\in \mathbb{R}^{C_m\times C}$ and $W_v\in\mathbb{R}^{C_m\times C}$ are shared learnable linear parameters. ``$\otimes$" represents the matrix multiplication.

We compute the products of $Q$ with ${K^G}^T$, and apply normalization and softmax function to obtain the attention matrix $A$, that is weights of $V^G$. For $a_i^j\in A$ is computed similarly to Eq.~\ref{eq:vanillaA}:
\begin{equation}
	a_i^j=softmax(\frac{\sum\limits_{c=1}^C{q_i^ck_i^{cj}}}{\sqrt{d_{s}}})
\end{equation}
where $q_i^c\in Q$, $k_i^{cj}\in {K^G}^T$. The letter ``$C$" means the number of the channel dimension. 

The attention matrix $A$ is multiplied by $V^G$ to obtain $Feats^G$. For $f_i^c\in Feats^G$, it can be expressed as follows:
\begin{equation}
	f_i^c=\sum_{s=1}^{S}{a_i^sv_i^{sc}}
\end{equation}
where $v_i^{sc}\in V^G$. The letter ``$C$" means the number of the channel dimension. The letter ``$S$" indicates the number of local $spots$.

The results of Local-Augment Attention are obtained by performing a linear transformation on $Feats^G$ to map its dimension to the original $spots^{m+1}$. Finally, based on Residual mechanism, the original $spots^{m+1}$ convey their features to these results to obtain the output of PLA. 

PLA concludes a further inference about $spots^{m+1}$ and emphasizes semantic information by utilizing their neighbors $spots^G\in spots^{m}$. 
\subsubsection{The architecture of PDMA}
\begin{figure}[h]%
	\centering
	\includegraphics[width =0.4\linewidth]{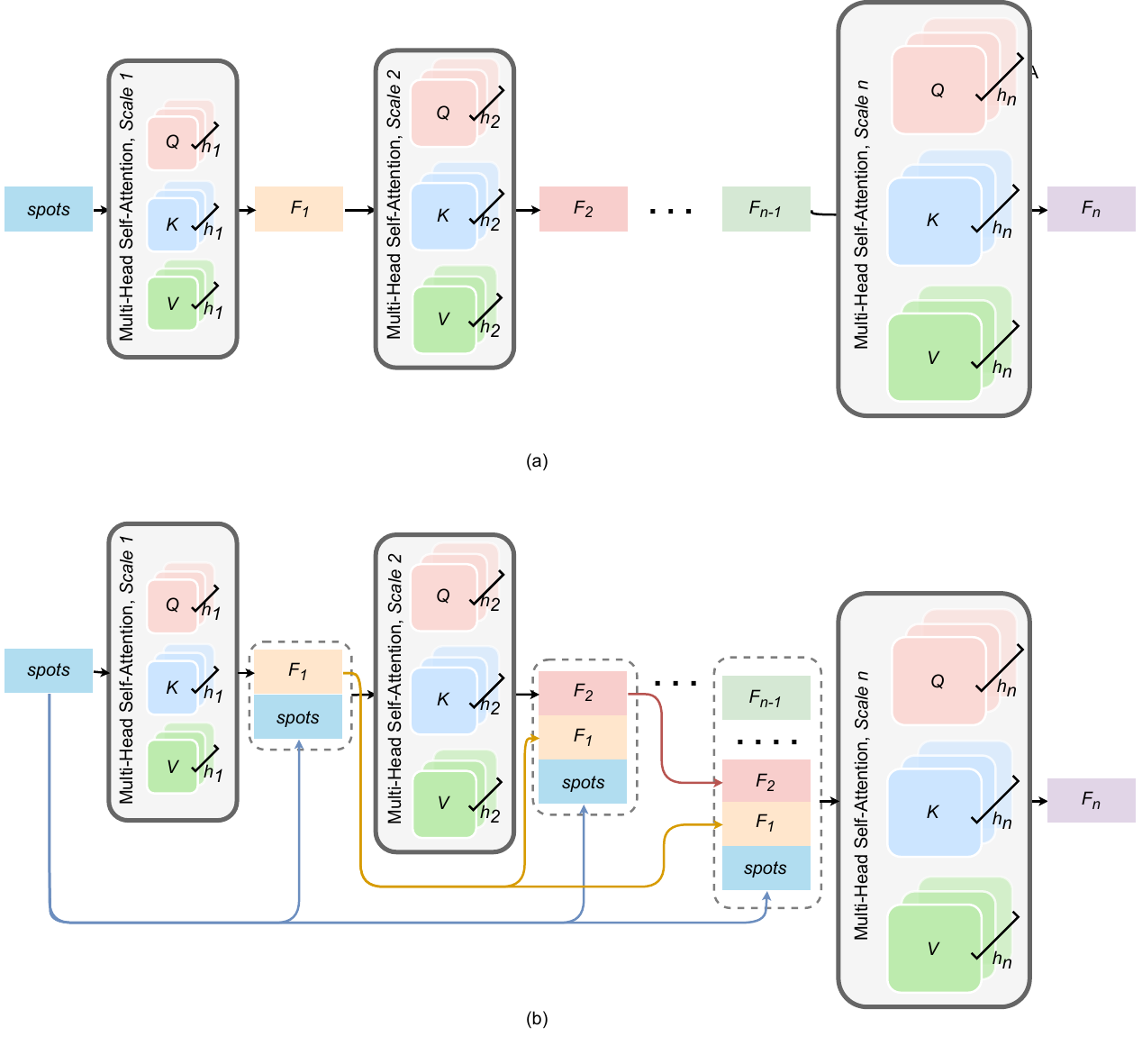}
	\caption{Visualization of the architecture of PDMA, (a) shows the multi-scale attention and (b) indicates the multi-scale attention with dense connection, $h_1=h_2=\ldots=h_n$.}\label{fig:PDMA}
\end{figure}
PLA promotes the inference of neighborhood information and updates each $spot$ by adaptively measuring the weights of the neighborhood features. However, different interactions of various $spots$ are not taken into account, which may not aggregate global information among $spots$ well. Therefore, it is necessary to explore the semantic information of $spots$. To explore relationships among these $spots$, we present Point Dense Multi-Scale Attention Module (PDMA) (Figure~\ref{fig:DRA} (d)). Although a vanilla  transformers\citep{vaswani2017attention} is competent at the task of extracting correlations of $spots$, our goal is to improve the representation of $spots$ by dynamically aggregating them depending on their semantic content. 

The vanilla transformers\citep{vaswani2017attention} has $h$ heads, and the shapes of $\{Q,K,V\}$ of all heads are the same. Different from the vanilla transformers\citep{vaswani2017attention}, our PDMA divides $h$ heads into $n$ groups, the shapes of $\{Q,K,V\}$ in each group are different. Figure~\ref{fig:PDMA} (a) visualizes shapes of $Q$, $K$ and $V$ of multi-scale attention. Output of the $n_{th}$ group with a scale of $sc_n$ can be learned from its previous group with a scale of $sc_{n-1}$. Eq~\ref{eq:vanillaSA} becomes:

\begin{equation}
	\label{eq:multiSA}
	F_{n}=\mathcal{H}(F_{n-1},W_q^{sc_{n}},W_k^{sc_{n}},W_v^{sc_{n}})
\end{equation}
where $F_{n-1}$ is the output of $(n-1)_{th}$ group.

We set the baseline of channel dimension as $C(C=64)$ and set the scale factor as $sc_n(sc_1=2,sc_2=4)$. The $\{Q,K,V\}$ with the scale $sc_n$ can be regarded as $\{Q^{sc_n},K^{sc_n},V^{sc_n}\}$. The representation of $Q^{sc_n}$, $K^{sc_n}$ and $V^{sc_n}$ of $n_{th}$ group can be changed from Eq.~\ref{eq:vanillaQKV} to:
\begin{equation}
	\label{eq:QKV}
	\begin{aligned}
		Q^{sc_n}&=F_{n-1}\odot W_q^{sc_n},\\
		K^{sc_n}&=F_{n-1}\odot W_k^{sc_n},\\
		V^{sc_n}&=F_{n-1}\odot W_v^{sc_n}
	\end{aligned}
\end{equation}
where $F_{n-1}$ is the output of the $(n-1)_{th}$ group of PDMA, $W_q^{sc_n}$, $W_k^{sc_n}$ and $W_v^{sc_n}$ are shared learnable linear parameters with the shape of $[C_{F_{n-1}}, C^{sc}]$, $C_{F_{n-1}}$ is the channel dimension of $F_{n-1}$, $C^{sc_n}=C\times sc_n$. 

To make full use of the features as much as possible, we also add the dense connection\citep{huang2017densely,iandola2014densenet,liu2019densepoint}. As shown in Figure~\ref{fig:PDMA} (b),
For the $n_{th}$ group of PDMA, its input is the concatenation of outputs from all previous groups. Therefore, Eq.~\ref{eq:multiSA} can be modified as:
\begin{equation}
	\small
	\label{eq:multidenseSA}
	F_{n}=\mathcal{H}((F_0\cdot F_1\cdot \ldots \cdot F_{n-1}),W_q^{sc_{n}},W_k^{sc_{n}},W_v^{sc_{n}})
\end{equation}

Similarly, Eq.~\ref{eq:QKV} becomes:
\begin{equation}
	\begin{aligned}
		\small
		\label{eq:QKVdense}
		Q^{sc_n}&=(F_0\cdot F_1\cdot\ldots \cdot F_{n-1})W_q^{sc_n}\\
		K^{sc_n}&=(F_0\cdot F_1\cdot\ldots \cdot F_{n-1})W_k^{sc_n}\\
		V^{sc_n}&=(F_0\cdot F_1\cdot\ldots \cdot F_{n-1})W_v^{sc_n}
	\end{aligned}
\end{equation}
where $F_0$ is the original input $spots$, $F_i(i=1,\ldots,n-1)$ is the outputs of $n_{th}$ group of PDMA. ``$\cdot$" means the concatenation operation.

In the end, results of all groups are spliced on the channel dimension, and the final output is obtained from these results spliced through MLP. PDMA sequentially aggregates different semantic information extracted from each group. As the PDMA deepens, representation is mapped to feature spaces with higher dimensional but information in lower dimensions may be lost. The dense connection approach complements the low-dimensional semantic context, aiding in recovering the details of the complete point cloud. PDMA achieves to extract the correlation among $spots$.

\subsection{Multi-resolution Point Fusion Module}\label{sec:Methods/MPF}
\begin{figure}[h]%
	\centering
	\includegraphics[width = \linewidth]{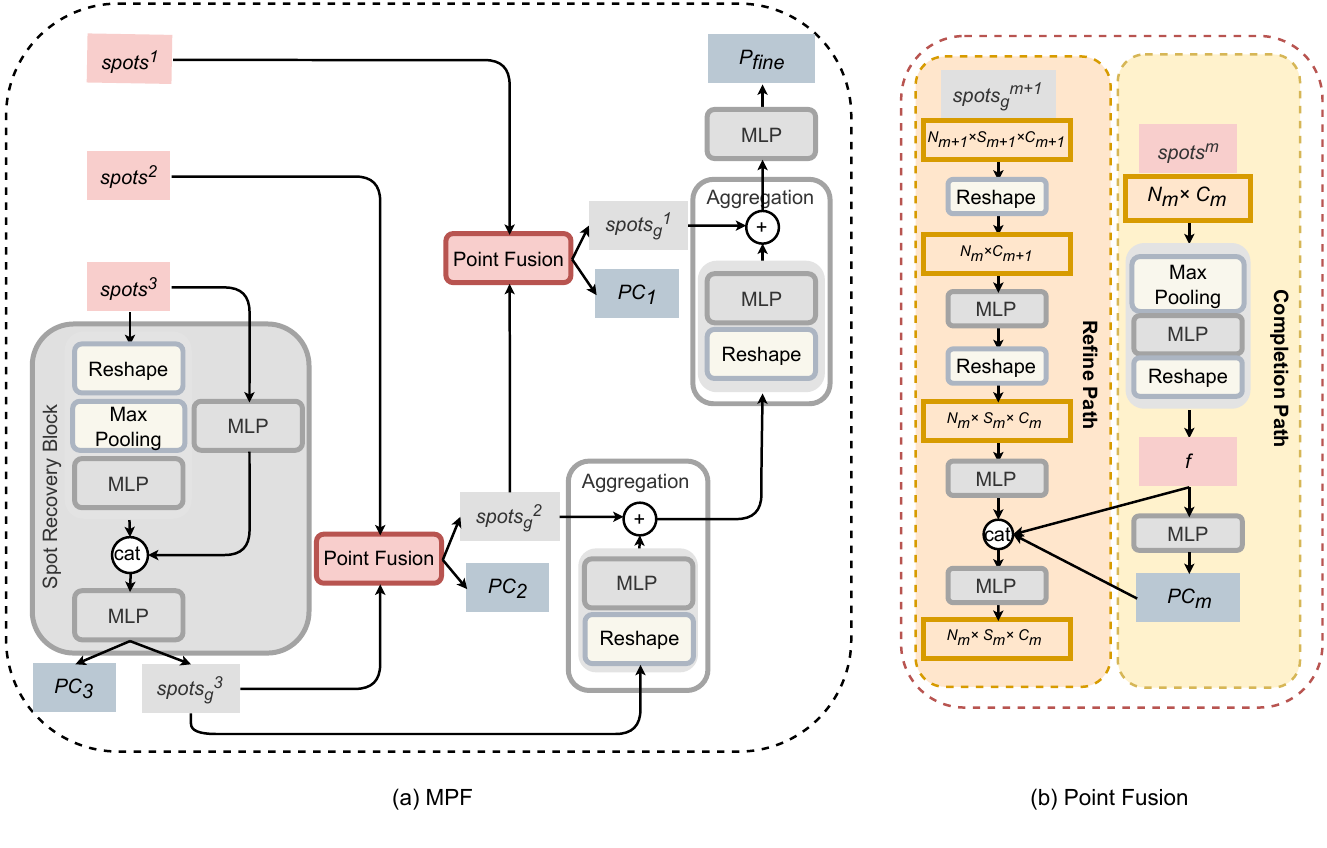}
	\caption{Illustration of the Multi-resolution Point Fusion Module and Point Fusion Block. The subscript ``$g$" indicates ``global".}\label{fig:MPF}
\end{figure}

According to these modules designed above, we gradually obtain $spots^m\left(m=1,2,3\right)$ from point clouds with different resolution. Each set of $spots$ has rich local information, and contains the global correlation. To recover a complete point cloud from these features, we propose the Multi-resolution Point Fusion Module (MPF) to complete a high-resolution point cloud. MPF first recovers \textit{global spots} to represent the information of the complete shape, and then uses \textit{spots} with different semantic contexts to generate complete point clouds with multiple resolutions. Each point cloud forcibly to map its spatial structure to the feature space to update \textit{global spots}. Therefore, under the restriction of the spatial location of the points, the distribution of the features matches that of the complete point cloud. For convenience of presentation, $spots^m_{g}$ denotes  \textit{global spots} in $m_{th}$ stage of MPF.

As shown in Figure~\ref{fig:MPF} (a), we follow the principle “from coarse to fine” to generate a complete shape. To begin with, $PC_3\in\mathbb{R}^{N_3\times3}$ and $spots^3_{g}\in\mathbb{R}^{N_3\times S_3\times C_3}$ are obtained from $spots^3$ by a Spot Recovery Block. $PC_3$ can be regards as skeleton points for grasping the complete shape and $spots^3_{g}$ represents the global features related to $PC_3$. Then $spots^3_{g}$ and $spots^2$ are aggregated to generate $spots^2_{g}$ with Point Fusion Block. Following the same method, the $spots^1_{g}$ is also obtained by $spots^2_{g}$ and $spots^1$. The $spots^3_{g}$, $spots^2_{g}$, $spots^1_{g}$ are aggregated in sequence to obtain point cloud $P_{fine}$ with the Aggregation block.

As shown in Figure~\ref{fig:MPF} (b), Point Fusion Block, as the key component of MPF, is denoted as the complete path and the refine path. The goal of the complete path is to generate a point cloud, and that of the refine path is to augment the reasoning of global features. In the complete path, $spots^m\in\mathbb{R}^{N_m\times C_{m}}\left(m=1,2\right)$ are used to generate $PC_{m}\in\mathbb{R}^{N_{m}\times3}$. In the refine path, $spots^{m+1}_{g}\in\mathbb{R}^{N_{m+1}\times S_{m+1}\times C_{m+1}}$ represent $N_{m+1}$ global features with $S_{m+1}$ local features. In order to refine these features, we reshape $spots^{m+1}_{g}$, and its shape becomes $[N_m,C_{m+1}](N_m=N_{m+1}\times S_{m+1})$. Then applying MLP and reshaping on it obtains $S_m (S_m={N_m}/N_{m+1})$ neighborhood information. In the end, $spots^{m}_g$ can be used to update itself by splicing intermediate features $f$ from complete path and $PC_m$ in channel dimension and then maps its dimension to the shape $\left[N_m,S_{m},C_{m}\right]$. The formula is as follows:
\begin{equation}
	\small
	spots^m_g=\phi(\phi(\mathcal{r}(\phi(\mathcal{r}(spots^{m+1}_g))))\cdot f\cdot PC_m) 
\end{equation}
where the letter ``$\phi$" means MLP, the letter ``$\mathcal{r}$" indicates reshape operation. The ``$\cdot$" is the symbol for concatenation.

\subsection{Optimization}\label{sec:Methods/Optimization}
Due to the unordered property of point cloud, it may be difficult to directly measure the distance between the generated point cloud and the ground truth. For this issue, \cite{fan2017point} propose Chamfer Distance (CD) and Earth Mover's Distance (EMD). We adopt Chamfer Distance as our loss function, because Chamfer Distance is more differentiable and more efficient than Earth Mover's Distance\citep{huang2020pf}. We define Chamfer Distance can be given as:

\begin{align}
	cd\left(P_c,P_g\right)=&\frac{1}{n_c}\sum_{c\in P_c}\mathop{\min}\limits_{g\in P_g}\parallel c-g\parallel_2+\nonumber\\
	&\frac{1}{n_g}\sum_{g\in P_g}\mathop{\min}\limits_{c\in P_c}\parallel g-c\parallel_2
\end{align}
where the symbol $P_c$ represents the completed point cloud with $n_c$ points, and the symbol $P_g$ represents the ground truth with $n_g$ points.

Our loss function consists of four items: $cd_1$, $cd_2$, $cd_3$, and $cd_{fine}$. It is formulated:
\begin{align}
	\mathcal{L}=&\alpha_1cd_1\left(P_1,P_g\right)+
	\alpha_2cd_2\left(P_2,P_g\right)+\nonumber\\
	&\alpha_3cd_3\left(P_1,P_g\right)+\alpha_{fine}cd_{fine}\left(P_{fine},P_g\right)
\end{align}
where $\alpha_1$, $\alpha_2$, $\alpha_3$, and $\alpha_{fine}$ are weights of out loss function. The value of $\alpha$ is illustrated in Section~\ref{sec:Experiment/Implementation Details}.

\section{Experiment}\label{sec:Experiment}
In this section, we first introduce datasets for point cloud completion and the evaluation metric. Then we show results of point cloud completion for our models and other methods.

\subsection{Dataset}\label{sec:Experiment/Dataset}
\textbf{PCN dataset} is one of the most common datasets for point cloud completion, which created from ShapeNet\citep{wu20153d} and published by PCN\citep{yuan2018pcn}. PCN dataset is composed of 30974 CAD models from 8 categories. The train data is obtained by rendering the 28974 CAD model from 8 angles. The test data is obtained by rendering the 1200 CAD model from only 1 angle. Back-projecting a 2.5D depth image into a 3D point cloud obtains the incomplete data. The ground truth contains 16384 points.

\textbf{MVP dataset} is a new dataset for point cloud completion, which is presented by VRCNet\citep{pan2021variational}. Point clouds of MVP dataset are sampled by Poisson Disk Sampling from 4000 CAD models grouped into 16 categories. Both train data and test data are obtained by uniformly setting 26 angles to renders these CAD models. Therefore, the number of train data and test data are $2400\times26$ and $1600\times26$.

\begin{table*}[h]
	\centering
	\caption{Results of point cloud completion on PCN dataset in terms of per point Chamfer Distance$\times10^4$(lower is better).}\label{tab:PCN}%
	\resizebox{1\columnwidth}{!}{
		\begin{tabular}{@{}lccccccccc@{}}
			\toprule
			Method & Airplane & Cabinet & Car   & Chair & Lamp  & Sofa  & Table & Watercraft & Avg. \\
			\midrule
			TopNet\citep{tchapmi2019topnet} & 2.25  & 5.21  & 3.66  & 6.13  & 7.27  & 6.86  & 4.60   & 4.59  & 5.07 \\
			PCN\citep{yuan2018pcn}   & 1.74  & 4.27  & 2.79  & 4.79  & 5.39  & 5.49  & 3.53  & 3.36  & 3.92 \\
			MSN\citep{liu2020morphing}   & 1.64  & 5.68  & 3.14  & 5.12  & 6.36  & 6.52  & 4.33  & 4.20   & 4.62 \\
			GRNet\citep{xie2020grnet} & 1.53  & \textbf{3.62} & 2.75  & 2.95  & 2.65  & \textbf{3.61} & 2.55  & 2.12  & 2.72 \\
			CompleteDT & \textbf{1.02} & 3.94  & \textbf{2.42} & \textbf{2.91} & \textbf{2.08} & 4.03  & \textbf{2.26} & \textbf{1.83} & \textbf{2.56} \\
			\bottomrule
		\end{tabular}
	}
\end{table*}

\begin{figure*}[h]%
	\centering
	\includegraphics[width = 0.75 \linewidth]{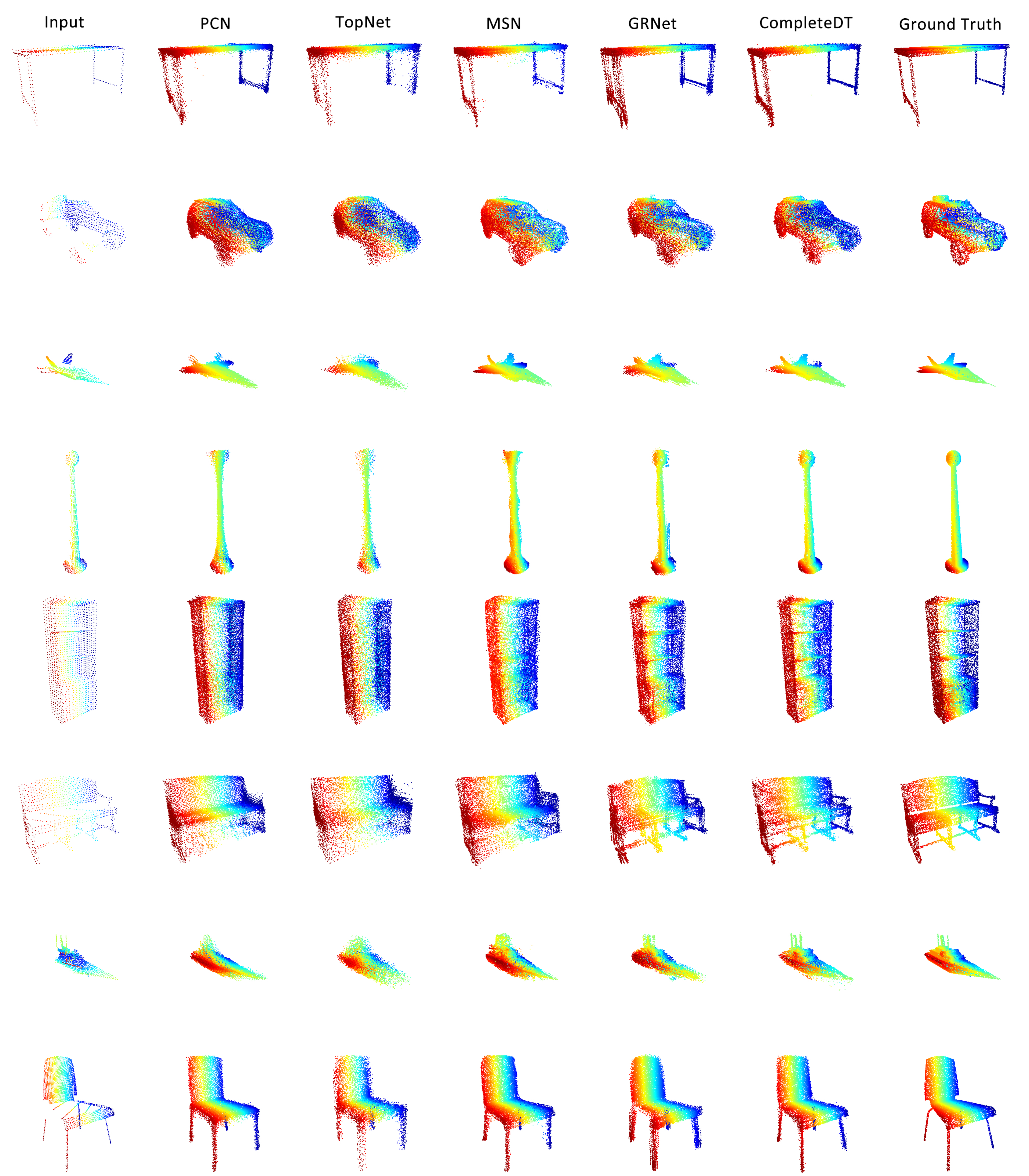}
	\caption{Visualization of point cloud completion on PCN dataset.}\label{fig:PCN}
\end{figure*}

\subsection{Implementation Details}\label{sec:Experiment/Implementation Details}
\textbf{Details for CompleteDT.} We obtain three multi-resolution point clouds $P_m$, whose sizes $N_m$ are 2048, 512 and 128, with IFPS. ResMLP mainly consists of the MLP Residual Block, having two MLP layers $[64-64]$. SGR gets $spots^m$ based on neighborhood points with SG. In Section~\ref{sec:Ablation Study/nsample}, we conduct an ablation study to explore the influence of the number of neighborhood points and decide to set the number to 48. In addition to SG, an MLP layer with the parameter of $C_{mlp}$ is needed to map the dimension of features of the low-resolution point cloud to that of the high-resolution point cloud. Then, the MLP Residual Block with four MLP layers $[(2\times C_{mlp}+2\times 3)-C_{out}-C_{out}-C_{out}]$ and MaxPool are performed. Finally, an MLP layer with the parameter of $\frac{3\times 16384}{N_m}$ is conducted. For three sampled point clouds $P_1$, $P_2$ and $P_3$, the features of $P_1$ is used to update that of $P_2$, and then the updated features of $P_2$ is responsible for the update of features of $P_3$. Thus, we use SGR twice, where parameters of $C_{mlp}$ are 64 and 128 in sequence, and parameters of MLP Residual Block are $[134-128-128-128]$ and $[262-256-256-256]$ sequentially. The important components of DRA are PLA and PDMA. PLA and PDMA receive $spots^m$. PLA learns local information adopting the attention mechanism, which has with four heads and each with a size of 64. PDMA consecutively uses multiple self-attention mechanism with four heads, and the dimension of each head is $64\times sc_i(sc_1=2,sc_2=4)$. Both PLA and PDMA obtain the outputs with the dimension $[N_m,C_m],C_m=\frac{3\times16384}{N_m}$. Then outputs are added together to update $spots$. Like SGR, DRA is used twice. Finally, MPF uses Spot Recovery Block to obtain $spots^3_g$ with the shape of $[128,1,384]$, and then uses the Point Fusion block to get $spots^2_g$ and $spots^1_g$, which shapes are $[512,4,96]$ and $[2048,8,24]$. Based on these $spots_g$, a complete point cloud with 16384 resolution is generated. For the values of $\alpha_1$, $\alpha_2$, $\alpha_3$ and $\alpha_{fine}$ in Section~\ref{sec:Methods/Optimization}, we follow the parameter configuration of VRCNet\citep{pan2021variational}, we set $\alpha_1$, $\alpha_2$ and $\alpha_3$ to 10, 0.5 and 0.5. For the first 5 epochs, 5-15 epochs and 15-30 epochs, we set $\alpha_{fine}$ to 0.01, 0.1 and 0.5. When epoch $>$ 30, $\alpha_{fine}=1$.

\textbf{Model configuration.} Our CompleteDT is implemented using PyTorch. We train our models using the Adam optimizer with initial learning rate $1e^{-4}$ (decayed by 0.7 every 40 epochs) and batch size 32 by RTX 3090 GPU. 

\subsection{Point Cloud Completion on PCN dataset}\label{sec:Experiment/PCN}
\begin{table*}[htbp]
	\centering
	\caption{Results of point cloud completion on MVP dataset in terms of per point Chamfer Distance$\times10^4$(lower is better).}\label{tab:MVP}%
	\resizebox{1\columnwidth}{!}{
		\begin{tabular}{@{}lccccccccccccccccc@{}}
			\toprule
			Method & \rotatebox{60}{Airplane} & \rotatebox{60}{Cabinet} & \rotatebox{60}{Car}   & \rotatebox{60}{Chair} & \rotatebox{60}{Lamp}  & \rotatebox{60}{Sofa}  & \rotatebox{60}{Table} & \rotatebox{60}{Watercraft} & \rotatebox{60}{Bed}   & \rotatebox{60}{Bench} & \rotatebox{60}{Bookshelf} & \rotatebox{60}{Bus} & \rotatebox{60}{Guitar} & \rotatebox{60}{Motorbike} & \rotatebox{60}{Pistol} & \rotatebox{60}{Skateboard} & Avg. 
			\\
			\midrule
			TopNet\citep{tchapmi2019topnet} & 2.65  & 4.44  & 3.68  & 8.87  & 14.14 & 5.74  & 7.04  & 6.05  & 10.42 & 5.62  & 7.33  & 2.75  & 1.45  & 4.27  & 4.77  & 2.54  & 5.74 
			\\
			PCN\citep{yuan2018pcn}   & 3.04  & 4.08  & 3.12  & 7.85  & 14.73 & 5.40   & 7.11  & 5.64  & 12.01 & 5.32  & 7.16  & 2.55  & 1.15  & 3.55  & 3.91  & 2.17  & 5.55 
			\\
			MSN\citep{liu2020morphing}   & 1.91  & 5.56  & 3.75  & 7.62  & 9.93  & 5.99  & 6.67  & 5.45  & 10.32 & 4.98  & 9.68  & 3.80   & 1.24  & 3.33  & 3.52  & 2.07  & 5.36 
			\\
			GRNet\citep{xie2020grnet} & 1.31  & 4.17  & 2.78  & 4.27  & 3.81  & 4.26  & 3.96  & 3.22  & 8.53  & 4.06  & 6.96  & 2.79  & 1.13  & 2.35  & 2.26  & 2.51  & 3.65 
			\\
			VRCNet\citep{pan2021variational} & 1.15  & \textbf{3.2} & \textbf{2.14} & 3.58  & 5.57  & 3.58  & 4.17  & \textbf{2.47} & 6.90   & 2.76  & \textbf{3.45} & \textbf{1.78} & \textbf{0.59} & \textbf{1.52} & 1.83  & 1.57  & 2.90
			\\
			CompleteDT & \textbf{1.08} & 3.65  & 2.35  & \textbf{3.55} & \textbf{3.60} & \textbf{3.51} & \textbf{3.44} & 2.54  & \textbf{6.00} & \textbf{2.50} & 3.97  & 1.88  & 0.61  & 1.71  & \textbf{1.59} & \textbf{1.07} & \textbf{2.69} 
			\\
			\bottomrule
		\end{tabular}%
	}
\end{table*}%
In this section, we compare the performance of our CompleteDT with other methods on the PCN dataset. We evaluate the results of point cloud completion in 16,384 resolution by adopting L2 metric of the chamfer distance. Compared with TopNet\citep{tchapmi2019topnet}, PCN\citep{yuan2018pcn}, MSN\citep{liu2020morphing} and GRNet\citep{xie2020grnet}, the performance of CompleteDT is improved by 2.51, 1.36, 2.06, and 0.16, respectively. Since the experiments in this section are based on 16384 points, we follow GRNet\citep{xie2020grnet} to combine the outputs of 2 times forward propagation to generate the point cloud with 16384 points. Table~\ref{tab:PCN} proves that CompleteDT is superior to the other methods, and has a powerful ability to infer the complete shapes from incomplete shapes. The shapes shown in Figure~\ref{fig:PCN} contain all categories of PCN dataset. We notice that some methods such as PCN,\citep{yuan2018pcn} TopNet\citep{tchapmi2019topnet} and MSN\citep{liu2020morphing} introduces noise into the shape of Table  (the first row). For the Airplane (the third row), the shape generated by TopNet\citep{tchapmi2019topnet} is oversmoothed. Although PCN\citep{yuan2018pcn} and GRNet\citep{xie2020grnet} try to make details finer, they also form a point cloud while having noise. Contrary to these methods, each shape generated by our CompleteDT preserves the details of the input incomplete point cloud and infers the missing parts with rich information. For the shape of Watercraft, it clearly shows that our CompleteDT achieves the best performance in geometry and details compared to the over-smoothing caused by other models. Overall, these qualitative and quantitative results indicate that our CompleteDT generates better details of shapes (e.g., Watercrafts and Lamps) than the other methods.

\subsection{Point Cloud Completion on MVP dataset}\label{sec:Experiment/MVP}
\begin{figure*}[htpb]%
	\centering
	\includegraphics[width=0.7\textwidth]{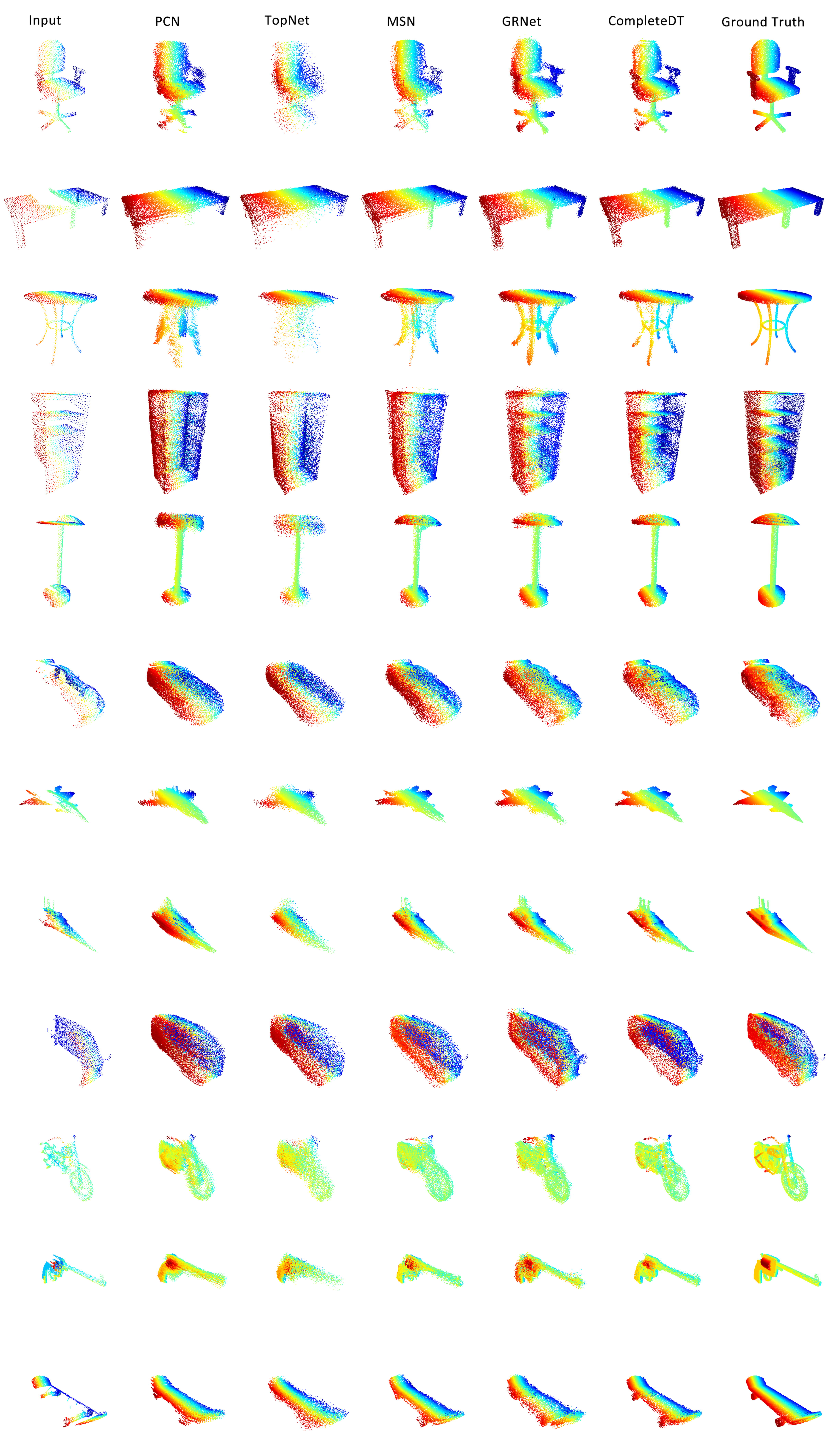}
	\caption{Visualization of point cloud completion on MVP dataset.}\label{fig:MVP}
\end{figure*}

In this section, we show the experimental results of different methods on MVP dataset. From the table, we achieve the best performance on most objects and get the lowest average value of 2.69 (multiplied by $10^4$). Compared to other methods, our performance improves by 3.05, 2.86, 2.67, 0.96 and 0.21 from Table~\ref{tab:MVP}. It is apparent from Figure~\ref{fig:MVP} that our method generates shapes with more details than other methods do. CompleteDT achieves outstanding results while retaining consistent geometries with existing parts. Despite the increasing number of categories, CompleteDT still performs satisfactorily and its generalization ability is equally good. For some categories such as Lamp, Chair and Table, CompleteDT produces reliable shapes. Like legs of the Chair in the third row, MSN\citep{liu2020morphing} tries its best to restore geometric details while generating a lot of noise. The point cloud generated by GRNet\citep{xie2020grnet} using 3D voxels as the intermediation is noise-free but denser than the ground truth. However, the point cloud produced by our model is much closer to the ground truth. 

\section{Ablation Study}\label{sec:Ablation Study}
We further analyze the proposed CompleteDT through ablation studies based on the evaluation on PCN dataset and use the chamfer distance as the evaluation metric. In this section, we discuss (1) the advantage of modules in proposed methods; (2) the effectiveness of PDMA; (3) the number of neighborhood points.

\subsection{The validity of modules in CompleteDT}\label{sec:Ablation Study/modules}
In this section, we investigate the validity of each module in the proposed CompleteDT. Recall that the CompleteDT is composed of four main modules: ResMLP, SGR, DRA and MPF. To verify the validity of above modules, we construct the following five networks. For the first network, it is used to assume that preprocessing operations can improve the performance of point cloud completion. It is composed of mentioned modules except ResMLP. The second network is used to prove the validity of SGR. Based on different resolution inputs, the SGR first obtains \textit{spots} and then updates them. To demonstrate the effectiveness of SGR for reasoning in \textit{spots}, we remove this module. Note that we temporarily supplement the acquisition process of \textit{spots} to DRA, since in the original network DRA processes \textit{spots} obtained from SGR. The third network contains mentioned modules except DRA. The fourth network is constructed by replacing MPF with MLP. The last network is the CompleteDT proposed. We evaluate the above networks on the PCN dataset with CD and their performance is listed in Table~\ref{tab:abModules}.
\begin{table}[htbp]
	\centering
	\caption{Results of ablation study about modules of CompleteDT in terms of per point Chamfer Distance$\times10^4$(lower is better).}\label{tab:abModules}%
	\resizebox{0.5\columnwidth}{!}{
		\begin{tabular}{cccccc}
			\toprule
			\multirow{2}[4]{*}{Network} & \multicolumn{4}{c}{Module} & \multicolumn{1}{c}{\multirow{2}[4]{*}{CD ($1\times 10^4$)}} \\
			\cline{2-5}    \multicolumn{1}{c}{} & ResMLP    & SGR   & DRA   & MPF   &  \\
			\midrule
			1st   &\ding{55}     & \checkmark    &\checkmark     & \checkmark    & 3.09 \\
			2nd   & \checkmark    & \ding{55}     & \checkmark     & \checkmark     & 4.42 \\
			3rd   & \checkmark     & \checkmark     & \ding{55}     & \checkmark     & 3.10 \\
			4th   & \checkmark     & \checkmark     & \checkmark     & \ding{55}     & 3.79 \\
			5th   & \checkmark     & \checkmark     & \checkmark     & \checkmark     & \textbf{2.56} \\
			\bottomrule
		\end{tabular}%
	}
\end{table}%
From table, our CompleteDT (the last network) largely outperforms other networks by 0.53, 1.86, 0.54 and 1.23. According to the result of the first network, it confirms our hypothesis that ResMLP can help subsequent modules to transform features and achieve improved performance. The second network also proves the effectiveness of the self-guided approach. However, only relying on SGR to extract features may not achieve the best performance, and the third network demonstrates that DRA addresses such limitation. For the fourth network, the significant increase in CD underlines the importance of MPF. Experiments show that removing any module can reduce performance and CompleteDT works quantitatively.

\subsection{The validity of DRA}\label{sec:Ablation Study/DRA}
\begin{figure}[htpb]%
	\centering
	\includegraphics[width=0.2\textwidth]{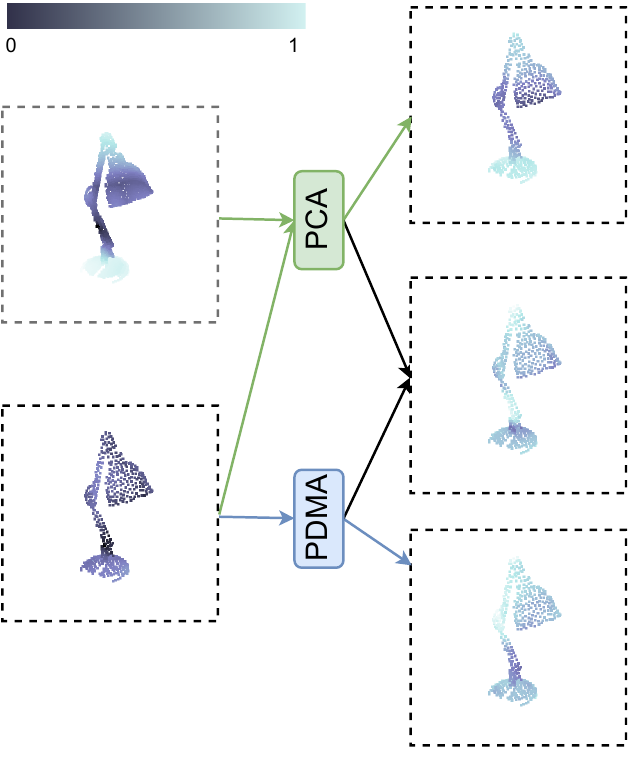}
	\caption{Visualization of feature weights mapped to the original point cloud.}\label{fig:abDRA}
\end{figure}

\begin{table}[htbp]
	\centering
	\caption{Results of ablation study about DRA in terms of per point Chamfer Distance$\times10^4$(lower is better)}\label{tab:abDRA}%
	\begin{tabular}{ccccc}
		\toprule
		\multirow{2}[2]{*}{PLA} & \multicolumn{3}{c}{PDMA} & \multicolumn{1}{c}{\multirow{2}[2]{*}{CD ($1\times 10^4$)}} \\
		\cline{2-4}
		&\makecell[c]{Common\\transformer} & No dense &Ours &  \\
		\midrule
		\checkmark     &       & \multicolumn{1}{c}{} &       & 3.00 \\
		\multicolumn{1}{c}{} &       & \multicolumn{1}{c}{} & \multicolumn{1}{c}{\checkmark} & 2.83 \\
		\checkmark     &       & \multicolumn{1}{c}{} & \multicolumn{1}{c}{\checkmark} & \textbf{2.56} \\
		\checkmark     & \multicolumn{1}{c}{\checkmark} & \multicolumn{1}{c}{} &       & 2.70 \\
		\checkmark     &       & \checkmark     &       & 2.65 \\
		\bottomrule
	\end{tabular}%
\end{table}%
In Section\ref{sec:Ablation Study/modules}, we discussed the undesirable performance without DRA, now we further analyze components of DRA. DRA consisting of PLA and PDMA is an important module to augment feature extraction. While completing the point cloud benefits from either PLA or PDMA, the combination between them is much important. Here we investigate whether combining them results in improved performance. As shown in Table~\ref{tab:abDRA}, combining PLA and PDMA results in a modest boost in the performance of point cloud completion, while using PLA or PDMA individually suffers a performance drop of 0.43 or 0.26 (the first two rows). Figure~\ref{fig:abDRA} indicates the visualization of feature weights mapped to the original point cloud. The lighter the color, the more important the learned feature is. The left of Figure~\ref{fig:abDRA} shows two point clouds of different resolutions from SGR, and the right of Figure~\ref{fig:abDRA} shows the visualization of features from PLA (the green arrow), PDMA (the blue arrow), and combined PLA and PDMA (the black arrow). It is apparent from Figure~\ref{fig:abDRA} that (1) PLA responds well to structures with obvious geometries, thanks to the enhanced inference of local features; (2) Different from PLA, PDMA provides opportunities for long-range interaction and therefore focuses on the global information. (3) The most striking result is that the combination of PLA and PDMA can both focus on global information and learn useful local features for geometric structures.

PDMA is one of the core modules of our CompleteDT and is well worth discussing deeply. Our PDMA processes features by taking the multi-scale attention allied with the dense connection. We provide the network where replacing PDMA with the common transformer\citep{vaswani2017attention} or multi-scale attention. From Table~\ref{tab:abDRA}, it is worth noting that while it performs 0.14 and 0.09 worse  than our CompleteDT, the multi-scale attention still leads to better results than the some SOTA methods like GRNet\citep{xie2020grnet}.

\subsection{The number of neighbors of the $spot$}\label{sec:Ablation Study/nsample}
After obtaining \textit{spots} containing local information based on neighbors between multi-resolution point clouds through SGR, PLA in DRA can update \textit{spots} within these neighbors, and PDMA in DRA provides opportunities for interaction among these \textit{spots}. Based on the above design, the number of neighbors is positively related to the performance and the computational cost of CompleteDT, so an appropriate number of neighbors is especially important for representing \textit{spots}. Specifically, a large number of neighbors may not improve performance, but may result in complex computations; only a few neighbors may fail to extract local features effectively resulting in poor performance. Thus, we conduct an ablation study on PCN dataset about the number $S$ of grouped neighborhood points, which $S$=16,24,48 and 96. As shown in Table~\ref{tab:abnample}, we compare the properties of models, like their sizes (number of parameters (Params)) and computational cost (floating point operations required (FLOPs)), on a single object.

\begin{table}[htbp]
	\centering
	\caption{Results of ablation study about the number $S$ of neighbors in terms of per point Chamfer Distance$\times10^4$(lower is better)}
	\begin{tabular}{c|c|c|c}
		\hline
		$S$ & Params (M) & FLOPs (G) & CD ($1\times 10^4$) \\
		\hline
		16    & 40.27 & 4.43  & 2.73
		\\
		24    & 40.27 & 5.30   & 2.63 
		\\
		48    & 40.27 & 7.90   & \textbf{2.56} 
		\\
		96    & 40.27 & 13.10  & 2.58 
		\\
		128   & 40.27 & 16.56 & 2.60 \\
		\hline
	\end{tabular}%
	\label{tab:abnample}%
\end{table}%
From Table~\ref{tab:abnample}, the chamfer distance of CompleteDT is better than that of other networks by 0.17, 0.07, 0.02 and 0.04, respectively. A small number of neighbors lead to an unsatisfactory expectation since missing local information, while a large number of neighbors result in increased computational cost. To balance computational cost and performance, we set the number of neighbors to 48.

\section{Conclusion}\label{sec:Conclusion}

In this paper, CompleteDT is designed to determine 3D geometric knowledge for completing point cloud. CompleteDT consists of ResMLP, SGR, DRA and MPF. ResMLP is a feature pre-extractor that lays the foundation for the performance of subsequent modules. SGR first collects local information to generate \textit{spots}, and then DRA is available for enhancing the reasoning of \textit{spots}. To be specific, DRA, a novel transformer-based module including PLA and PDMA, is proposed for augmenting the inference of features extraction. PLA facilitates the extraction of information with each \textit{spot}, and PDMA effectively provides global correlation between \textit{spots}. Finally, MPF generates high-quality complete point clouds in the way to update features with the help of points. Evaluations on PCN dataset and MVP dataset show that CompleteDT based on the above modules gets more expressive features to complete shapes and outperforms some state-of-the-art methods.

\bibliographystyle{unsrtnat}
\bibliography{references}  






\end{document}